# Limitations of Deep Neural Networks:
# a discussion of G. Marcus' critical appraisal of deep learning

Stefanos Tsimenidis

**Abstract**

Deep neural networks have triggered a revolution in artificial intelligence, having been applied with great results in medical imaging, semi-autonomous vehicles, e-commerce, genetics research, speech recognition, particle physics, experimental art, economic forecasting, environmental science, industrial manufacturing, and a wide variety of applications in nearly every field. This sudden success, though, may have intoxicated the research community and blinded them to the potential pitfalls of assigning deep learning a higher status than warranted. Also, research directed at alleviating the weaknesses of deep learning may seem less attractive to scientists and engineers, who focus on the low-hanging fruit of finding more and more applications for deep learning models, thus letting short-term benefits hamper long-term scientific progress. Gary Marcus wrote a paper entitled *Deep Learning: A Critical Appraisal,* and here we discuss Marcus' core ideas, as well as attempt a general assessment of the subject. This study examines some of the limitations of deep neural networks, with the intention of pointing towards potential paths for future research, and of clearing up some metaphysical misconceptions, held by numerous researchers, that may misdirect them.

## I Introduction

Pattern classification with neural networks is nothing new, first proposed in 1947 [1] and implemented during 1957-58 by Rosenblat [2]. Although Minsky and Papert [3] brought to the fore the limitations of these perceptrons and the problem of linear separability, famously depicted in the XOR problem, it was always known that these limitations could, in principle, be surpassed by adding hidden layers. The issues that hampered the real-world implementation of such deep neural networks were solved much later, through (1) the re-invention of back-propagation [4], (2) the running of deep neural networks on GPUs, allowing for the scalability of the computationally expensive matrix calculations needed, and (3) the collection of massive amounts of labeled data, indispensable for training deep neural networks [5].

Deep learning as we know it took off around 2010 [6] through the work of researchers Geoffrey Hinton, Yoshua Bengio, Yann LeCun, and the IDSIA group in Switzerland, and the tipping point came in 2012 with a series of publications showing how deep architectures were increasingly achieving state-of-the-art performance in a variety of pattern classification tasks, most notably the famous work of Hinton and colleagues on the ImageNet object recognition challenge [7]. This sparked a new machine learning revolution, with an explosion of applications in technology, science, finance, industry, even art and education. The previous AI winter gave way to a surge of activity and excitement, and in direct proportion with the media over-hype surrounding "artificial intelligence," the expectations of the public skyrocketed. People extrapolated into the future a wide range of fantastical scenarios, from the Promethean storyline where AI helps humans achieve god-like status[8], to the dystopian and paranoid themes of AI enslaving or destroying us[9].

In the present day, almost a decade after the "deep learning revolution", what previously appeared like an exponential curve of progress towards AGI (the mythical Artificial General Intelligence) starts to resemble a sigmoid, an exponential that reaches a plateau. After the advent of cars that stay inside highway lanes absent human steering, programs that beat humans at chess, Go and video games, and chatbots with storytelling skills and word-play capabilities, everyone braced oneself: something big was about to come. Now, self-driving cars haven't gotten any nearer full autonomy in an urban environment, game-playing programs keep playing games, and chatbot applications are being shut down after failing to rise to the expectations. Some of the pioneers of the Deep Learning revolution are now worried we're about to enter a new AI winter [6] [10].

What follows is a discussion of some of the most severe limitations of deep neural networks, as identifying a problem is often the first step towards solving it.

**II Deep Neural Networks**

Neural networks, named as such because they loosely model the computational operations of biological neurons, consist of input units, one or more hidden layers of processing units, and a set of output units. These artificial neurons are interconnected, and information flows from the input units forward, through the hidden layers, and finally to the output.

What the network does is it maps the inputs to the desired outputs. The information for this mapping is represented in the neuronal connection weights, which determine what computations will be performed on the input signal. Training the neural network entails adjusting these weights, using the back-propagation algorithm, to gradually make a structure that will process the inputs and get them to approximate the desired outputs.

With enough training examples neural networks can, in principle, approximate any function, that is, any possible input-output mapping. In practice they never reach perfect generalization, but they often perform well enough for a large number of narrow applications, hence their rising popularity.

When the number of hidden layers increases we get "deep" neural networks, which have impressive capability in learning and representing input-output mappings.

Deep neural networks are mostly used for classification tasks, assigning an input into a particular class from a set of possible classes. Many applications, from computer vision to machine translation, can be formulated as classification problems. In a sense, neural networks are used like a hammer in quest for nails: machine learning engineers are on a constant lookout for tasks that can be expressed as a class-assignment problem.

In principle deep neural networks can approximate every conceivable input-output mapping, and in principle a huge amount of cognitive tasks can potentially be formulated as a classification problem. In principle modeling the computations performed by biological neurons could give rise to intelligence, and deep neural networks could, in principle, be the final invention humanity had to make.

**III Limitations**

Intelligence, according to connectionism, emerges from computations in neurons and networks of neurons, and since humans exhibit high levels of intelligence, the connectionist thesis is supposedly proven right. What follows from this is that the limitations, if any, of deep neural networks, should be minor and unimportant. These digital models of brain function are still crude and primitive, connectionism tells us, and with further progress in the field, with deeper and more complex architectures, their shortcomings should gradually diminish to zero, at which point we will have reached the "AI Singularity" [11] [12].

The empirical data shows that the limitations of deep neural nets are neither unimportant nor diminishing over time.

**Generalization**

Neural networks are infamous for their inability to generalize. Given plenty and good training data, and with hyperparameter calibration, they may yield impressive results in interpolation, but in terms of extrapolation their failure is absolute.

For humans generalization, transfer learning, and extrapolation comes easily and naturally. As Fodor and Pylyshyn [13] say, "the ability to entertain a given thought implies the ability to entertain thoughts with semantically related contents." Any effort towards artificial intelligence and applying machine learning to real-world situations must take generalization seriously into account.

All machine learning models struggle to make the best tradeoff between approximation and generalization [14]. Approximating the training data too closely leads to overfitting, and settling for a loose fit for the sake of generalization leads to an unacceptably bad performance on the training data. In most practical applications, the training data represent the cases most commonly encountered. On an intuitive level, focusing on the common cases makes for a model that fails spectacularly on unexpected situations, and crafting a more flexible model renders it useless on the everyday, normal cases that the model was meant to tackle for in the first place.

There are good reasons why metrics, confusion matrices, ROC curves, and MSEs are used, and why we have the generalization\overfitting tradeoff and the bias\variance tradeoff [14] [15] [16]. Automating classification tasks will always entail tradeoffs.

Unfortunately, in high-stake and in critical situations tradeoffs are unacceptable.

When human lives are at risk, or when a wrong decision will have dire consequences, we encounter a paradox. In these high-stake situations we want our model both to generalize perfectly, but also to perfectly overfit the training data. You want a self driving car to both adapt to new situations, which means underfitting the training data, but also to always hit the brakes when pedestrians pass the street, which necessarily means overfitting the training data.

As the industry is conspicuously silent on this point, perhaps it merits emphasizing. Let alone the failure of deep neural networks to extrapolate, even interpolation poses problems for them. Either they will perform well in general cases and fail in edge cases, or they will overfit the training data and fail everywhere else. For certain applications this trade-off is prohibitive. No one wants to pass a street roamed by self-driving cars with a 0,87 probability of hitting the brakes. A car biased towards false negatives will be deadly, while a car biased towards false positives will be useless.

A direct consequence of these issues is that you can not use neural networks to make predictions about policies and economic measures. Just like they don not know what to do with data they've never seen before, neural networks can't predict the outcomes of new, never-before-implemented, policies.

It has been argued [17] that the backpropagation learning algorithm may be responsible for the inability of deep neural networks to generalize well. The output units are independent and compute error functions independently of each other, the argument goes, while a more "global" type of learning would allow the model to encode relationships among features. It does not seem obvious, though, how a more globalized training regime would solve generalization and succeed where every single machine learning algorithm ever devised has failed.

**Data Dependent**

Deep neural networks are extremely data-dependent in that they require massive amounts of labeled data to learn adequately [5].

It has been known for decades that neural networks can not generalize well and can not extrapolate at all [17]. The only thing that saves the day is exhaustive data, teaching them to approximate a function with many, and in theory infinite, instances. That is why deep learning took off during the past decade: massive amounts of labeled data, and the newfound ability to harvest them, that is, more computational power to deal with them, and backpropagation to learn from them. In some sense, the modern machine learning revolution is more due to big data than it is to neural networks.

We can not use deep learning in situations where there are not massive amounts of data in the first place. First we must generate and label a large volume of data, then use a deep learning model to memorize them. In any new application, in any uncharted territory, humans must go first, map the territory, then train a model.

Humans can learn from a few trials, even in a single trial (one-shot learning), and generalize easily from that one experience. A lot of hype surrounds some accomplishments of "AI" algorithms, with the implied thesis that these algorithms are somehow "intelligent," but the truth is these models simply memorize billions of data, interpolate at the short gaps between, and thus might appear to "understand." They yield seemingly impressive results only when they've already been fed almost every possible input\output pair [18].

Backpropagation, one of the great deep learning enablers, needs labeled data to compute error functions and adjust synaptic weights. Without knowing the outputs there can be no learning--a vicious circle if you lack big data and human annotators.

**Can not Function in an Open World**

Machine extrapolation is impossible because machines must generalize from finite data, and unless the function to be approximated is very simple, or periodic, they are bound to fail outside of the subspace enclosing the instances they have encountered. This doesn not necessarily have to pose a problem in finite environments, and given big data that span all the input space, in which case the neural network only needs to interpolate. Unfortunately, systems in the real world are not isolated. They are complex and open-ended. And deep learning models will have to extrapolate.

A divide-and-conquer strategy could be attempted to break the domain into smaller problems and train a separate neural network for each, but most real world problems are not easily compartmentalized, especially in fields like policy-making and economics. In these highly open-ended systems the number of variables and factors affecting the outcome is vast, factors that are in turn affected by other factors, on and on, a chain of causal relationships reaching to infinity [19].

In such a system no amount of training data will be adequate as it would only broaden the range of interpolation. An infinite problem space can not be approximated by the finite subspace of training instances, and this is true even in the era of big data.

## Can not Adapt

For the same reason that neural networks ca not function in an open world, they also can not adapt to outlier inputs in closed systems. Exhaustive data that cover all possible situations are feasible not in finite worlds, but in worlds that are both finite and relatively static.

Deep learning models fail in dynamical systems. As the environment changes, they must be trained anew. In "black swan" situations [20] they fail completely, and should not be utilized in environments where such occurrences are expected.

## Lack of Hierarchical Perception

A sentence is not a series of words, but a series of phrases, each one with a unified meaning [20]. A picture is not a series of pixels, but an arrangement of objects, some of which are arrangements of objects themselves (eg a car). Humans naturally discern such hierarchies and process them as unified wholes, variables in relationship with other variables, forming hierarchies in relationship with other hierarchies. Artificial neural networks can not detect such structures [22]. Their attention is uniform; every element of their inputs, every pixel of an image, every word of a sentence, carries equal weight.

According to Marcus [17], neural networks "do not include any explicit representation of a relationship between variables. Instead, the mapping between input and output is represented through the set of connection weights....They replace operations that work over variables with local learning, changing connections between individual nodes without using 'global information.'"

Their oblivion to linguistic structure is dwarfed by their incomprehension of abstract relationships. The phrase "John is father to Mary" is meaningless to them and they can not register the conceptual similarity with the phrase "George is father to John." As known since the 1960's, though, neural networks could appear to understand abstract relations "if they're trained by an exhaustive procedure of taking in every possible datum where the relationship holds," (Rosenblatt (1962) ) in other words, by being spoon-fed the entire input space, which is not unlike what takes place today with the highly publicized and much hyped "accomplishments" of modern "AI."

## Causation vs Correlation

Similar to their inability to register hierarchical structure and abstract relationships, neural networks lack the capacity to differentiate between correlation and causation [23]. Their ability to spot factors that correlate highly with each other within stupendous volumes of complex data is truly remarkable, and can be harnessed for many practical applications but inferring causal relationships eludes them.

This may lead to "dumb insights," especially in cases of overfitting, where they're prone to confuse coincidences for meaningful associations (a well known example being the Google flu case [24]). The capacity of "AI" models, which usually means deep learning models, to attribute statistical significance to spurious correlations has gained an almost legendary status.

The inability of neural nets to detect causation heavily aggravates their inability to make predictions. A neural network only perceives a graph of points on an axis system. Only a human can "see behind" these points to what, and how, variables affect each other, and thus is able to anticipate events. Deep learning models can make adequate short term predictions in simple systems or with exhaustive training, but in a complex system a capacity for causal inference is indispensable.

Work on mathematically formulating causal inference has been attempted [25], but not with deep learning. The causal-inference research project has focused on higher level algorithms, on cognitive models instead of connectionist ones. It seems that if the task of automated causal reasoning is ever solved it will not be with neural networks.

**Blackboxes**

In engineering a black box signifies a system or device we can not know how it functions but can only see what inputs go in, and what outputs come out. In deep learning we use feature extraction and vectorization to represent the objects we want the model to process with numbers. The model only sees numbers and spots statistical regularities among these numbers [6]. It can not register any qualitative relationships between the variables these numbers represent, like causality, hierarchy, and other abstractions [17] [26]. It only detects quantitative relationships among the numbers themselves, and therefore cannot explain its decisions in any human-meaningful way. It's a black-box.

It has been shown that a machine learning model's interpretability is inversely proportional to its flexibility [16], and neural networks, with their mimicking brain plasticity, are arguably the most flexible models of all [27].

Debugging such an algorithm poses serious problems. Many applications use cascades of deep neural networks, one's output feeding the input of another to achieve complex tasks. The human brain is not merely a large neural network but it is a network of networks, and the quest for artificial general intelligence may take the direction of researching hierarchies of deep neural networks [28]. These systems might well be impossible to debug.

As long as they remain blackboxes we can't trust neural networks to make important decisions in high-stake situations [26]. Netflix recommendations and automatic captioning of photos on blogs might be fine, but terrorism detection and forensic procedures shouldn't be entrusted on systems that can not explain how they reach their conclusions.

In Lipton [29] we learn that concerns over trust and other issues of interpretability may be "quasi-scientific." We learn that a lot of disagreement has been going about what makes a model interpretable, with candidate definitions often contradicting each other, therefore, we learn, concerns about interpretability are meaningless and "quasi-scientific".

While it is true that interpretability is not always well-defined and there is high-variance of competing definitions, all of these definitions share one thing in common. Neural networks do not fulfill any one of them. No matter what standards for interpretability you set, neural networks do not meet them.

When and if deep learning models start becoming easier to understand the time will be ripe for a more rigorous examination of the issue, and the "quasi-scientific" discussions about AI explainability will be a thing of the past.

**Ontological Inference**

Among the abilities to understand hierarchical structure, causal relationships, abstract ideas, and symbols, ontological inference may be the highest abstraction of all, and to the author's knowledge, has never been discussed in the artificial intelligence literature.

Ontological inference is the capacity to infer, from sense perceptions, the existence of sense-undetectable entities. The proverbial apple falling on a robot's head will never cause it to discover gravity. Though the robot may register statistical regularities about objects falling towards the ground when unsupported, it will never hypothesize an entity, an existence, that actively pulls bodies towards the center of the planet. Atoms and electrons have never been and will probably never be directly observed, but their existence can be inferred from other observations, indirectly [30]. Even a natural process can be, and is, conceptualized as an entity, an abstract being of sorts, and its occurrence can be expected whenever the conditions that trigger it are present.

This ontological-detection faculty that humans posses is perhaps the single most important factor that enabled humanity's scientific progress and flourishing. To say that deep neural networks don't have access to this higher function would be an understatement. Ontological inference is so outside the grasp of any mathematical formulation and computational model that it is never being discussed in the literature of any modern, reductionism-infused science that supposedly studies the human mind: cognitive science, neuroscience, psychology, computer science, even modern philosophy and epistemology. Everyone seems meticulous and rigorous in pretending that (1) such a faculty does not exist, and (2) we are well on the way to create machine intelligence that far outweighs the intelligence of humans.

**Technical Debt**

The term "technical debt" refers to systems that seem attractive on the short-term but become unmanageable later on when they increase in complexity.

Traditional systems using rules and logic can be modularized, debugged, and polished. As a system increases in complexity such tasks become increasingly important, not because they add further functionality, but because they clear the path for future improvements.

The reason why deep learning is utilized, though, is precisely because it excels in tasks where more logic-oriented techniques fail. Deep learning models may be easy to implement and may bring some impressive initial results to companies, but as they scale up, and since they lack the debugability of interpretable algorithms, small hindrances can compound to serious issues.

Also, their penchant to mistake correlation for causation rises proportionally with the increase in all the data, variables, features, and quantities that the neural networks are summoned to inspect, until the noise overwhelms the signal.

**Sheer Lack of Intelligence**

Lately, a few popular treatments of artificial intelligence have commented of how deep neural networks can produce outputs that seem, from a human perspective, monumentally unintelligent [31] [32]. Algorithms that missclasify TV sets for orangutans, generate dirty words and racist remarks when asked to generate children's stories, give ridiculous answers to simple common-sense reasoning questions. None of that comes as a surprise to those who know how these algorithms work, but they serve as a reminder that the intelligence we're sometimes carried away to attribute to them is an illussion.

A lot has been said about games-playing AI programs that use deep neural networks trained with reinforcement learning [33]. They beat humans and come up with seemingly ingenious and creative solutions to problems but, again, to attribute some sort of understanding to these algorithms would be premature. DeepMind's Atari system failed on minor perturbations from the training set, such as slightly moving the Y coordinate of an object [34]. These neural-nets were trained day and night for weeks, on more than a thousand processors and GPUs, using petabytes and petabytes of training data, then playing games against each other to further "hone their skills," eventually achieving breathtaking results; and then a minor rule changes slightly, or a hardly perceptible perturbation is applied on the pixels, and everything crumbles. The algorithms never really understood anything about what they were doing, but simply overfitted the data and superficially acted upon arrays of pixels without awareness of what these pixels represented.

Another famous case is the adversarial attacks on deep neural networks trained for object recognition from images [35] [36]. These networks can be easily tricked into misclassifying images by applying slight perturbations, which can be found by maximizing their prediction error [37]. No neurologically sound human would misclassify an image of an ocean for a car. Artificial neural networks simply don't perceive things the way we do.

## IV The Rise and Fall of Connectionism

The lure of connectionism initially came from cognitive science and the doctrine that all parts of intelligence, including abstract thinking and symbol manipulation can be accounted for by the computations of networks of neurons [38].

This view initially seemed to be corroborated by the apparent success of deep neural networks, but has lately been challenged. There's growing evidence from neuroscience that knowledge of brains does not correlate with understanding of behavior [39]. Even if we simulated the entirety of a human brain there are no guarantees, to put it mildly, that the simulation would do what we would expect a human brain would do. The connectionist approach to general intelligence is being abandoned.

Some final attempts to defend connectionism point to it as the biologically plausible way of explaining and realizing intelligence, but back-propagation, the fundamental learning algorithm that largely made connectionism implementable in the first place, is clearly not biologically plausible [40], and the parts of deep neural networks that are biologically plausible have severe limitations.

New approaches are being proposed, symbolic representations [22], algorithmic representational models [41], conceptual representations of higher cognitive functions [12], a combination of approaches [42], etc. All these proposals attempt a higher-level approach to intelligence than the neural, as scientists increasingly acknowledge connectionism is not the answer. Reproducing brain computation does not bring us nearly as close to general intelligence as previously hoped.

Deep learning has tremendous potential for practical applications, most of it still untapped, and with the late massive inflow of talented engineers into the field the future looks promising. If we ever succeed in creating artificial general intelligence deep neural networks will no doubt have played a role in it, however small that role might be.

## V Conclusion

Ever since Searle made his Chinese room argument [43] a lot has been said to refute it [44] [45] [46], even to the the point of mocking Searle for having made it. Perhaps Searle gets the last laugh, though; nearly all the limitations of deep neural networks we examined, as well as the fact that these limitations are neither trivial nor diminishing with new breakthroughs in computer science, are because Searle was right.

Neural networks do not understand what they do.

They may be able to represent information, but they have no awareness of what this information means. It remains to be discerned to what degree is intelligence dependent on, and rendered possible by, consciousness.

**Supplementary Materials**

**Report on the current state of AI explainability**

Since lack of explainability is a big weakness of deep learning, and discussed in the main body body of this paper, we include a short report that was commissioned for a grant proposal, for funding to conduct research on an experimental, explainable machine learning model. The constraints were: a) the report should be about a page long, and b) it should contain references only from 2019, since the grant proposal would be sent in the first month of 2020, and we needed to capture the current state of the field. The reference-numbers do not point to the references of the main paper, but to the references at the end of this document.

De Graaf and Malle [1] argue that humans regard AI systems as intentional agents and expect explanations similar to what a human would give, thus linguistic and psychological analysis of human-level explanations is necessary. Rutjes et al [2] maintain that the problem of AI interpretability lies outside the domain of computer science and should be tackled by cognitive scientists. They even suggest seeking counsel from the social sciences, since concepts like fairness, accuracy, and explainability may be, according to the authors, social constructs. Much focus has been given on these ideas and similar arguments have been made. These are important and nuanced issues, and when black-box models cease being black-boxes, these discussions will become relevant.

The class of inherently explainable machine learning algorithms is comprised by linear models, decision trees, and Bayesian classifiers [3]. Explanations entail providing visualizations and information about coefficient weights, decision paths, rules and probabilities. In one study with Logistic Regression predicting diabetic patient re-admittance [4], the model displays probabilities of re-admittance, a confusion matrix with metrics, and simple statements such as "The probability of readmission increases as the patient's glucose serum levels exceed >200." Another study uses gradient-boosted trees for diagnosis and applies techniques to explore counterfactuals, identify key factors, and display visualizations [5]. With these interpretable models the cognitive and linguistic analysis of explanations mentioned above is not irrelevant. Hal et al [6] developed a systematic interview method to identify the explainability requirements of an AI system's stakeholders and thus be able to design a system that meets them.

At the other end of the explainability spectrum lie Deep Learning models that are, by their very nature, blackboxes. Efforts to render them intelligible utilize tools to approximate deep neural networks with either linear or gradient-based models, or decision trees [3]. These techniques are effective only for a narrow domain of the model they approximate, and break down outside of it [7]. An example of this in [8] focused on computer vision. The researchers performed correlation analysis to detect patterns between the factors of input data variations and test outputs. The algorithm learns a dictionary of semantic concepts as an explanation of what it "sees" in the image. A similar correlation analysis was used in a classification task (determine whether MRI scan requests should be approved) [9]. The goal was to find specific words in the MRI requests that maximize probability of positive prediction.

Despite such efforts deep neural networks refuse to yield. They remain blackboxes, and many researchers have given up attempts to unlock them, searching for workarounds instead. In [10] it was suggested that the model should augment classifications by providing additional information and predictions, thus appearing more rigorous and authoritative than it is (eg., a model that predicts the likelihood of future dementia could also predict the results of future cognitive tests and brain scans). Additionally, they suggest that reproducibility might make up for the lack of explainability: If a machine learning pipeline consistently yields high-quality results on a variety of datasets we feel justified to trust it.

Akula et al [11] experimented with an interactive game where a human sees a blurred image and, to infer what it depicts, asks the machine questions. The machine runs object recognition algorithms on the unblurred image. The user asks questions that are meaningful to a human, and the machine iteratively builds a model of what the human perceives\thinks, learning to provide human-relevant answers. Then the user hypothesizes which objects the machine will classify correctly in a set of unblurred images. The closer his expectations are to actual machine performance, the more the user is said to trust the machine. It's unclear what the purpose of such studies is, but experiments showed that over time humans trust the machine more, regardless of the neural network's remaining as much a blackbox as before.

Ehsan et al [12] worked on reinforcement learning, where future decisions depend on past ones. Users train the model by acting on the specific domain\task and providing natural language explanations of their decision-making rationale. The system associates action with rationale, and during deployment it outputs, for each action, the corresponding explanation. This pseudo-explainability method yields acceptable results only in narrow domains where the number of possible states and actions is severely limited, hence in the study the domain problem was a simple video game.

Such techniques that give the illusion of explainability may enhance the user's trust and comfort, but in a plethora of applications, especially high-stake ones, we need the explainability to be actual. When the scientific community gives up on the problem and experiments with workarounds, it is time to consider alternative AI models.